\documentclass[runningheads]{llncs}
\usepackage{graphicx}

\usepackage{tikz}
\usepackage{comment}
\usepackage{amsmath,amssymb} 
\usepackage{color}

\usepackage[accsupp]{axessibility}  

\usepackage{cite}
\usepackage{hyperref}
\usepackage{booktabs}
\usepackage{threeparttable}
\usepackage{multirow}
\usepackage{caption}
\usepackage{subcaption}
\usepackage{bm}
\usepackage{bbm}
\usepackage{siunitx}
\usepackage{orcidlink}

\hypersetup{colorlinks}

\DeclareMathOperator*{\argmin}{arg\,min}

\usepackage{xspace}
\makeatletter
\DeclareRobustCommand\onedot{\futurelet\@let@token\@onedot}
\def\@onedot{\ifx\@let@token.\else.\null\fi\xspace}

\def\etal{\emph{et al}\onedot}
\makeatother

\begin{document}
\pagestyle{headings}
\mainmatter

\title{Hunting Group Clues with Transformers for Social Group Activity Recognition}

\titlerunning{Hunting Group Clues with Transformers for SGAR}
\author{Masato Tamura\orcidlink{0000-0003-1029-5271} \and
Rahul Vishwakarma\orcidlink{0000-0001-8874-6816} \and
Ravigopal Vennelakanti}

\authorrunning{M. Tamura \etal}
\institute{Hitachi America, Ltd.\\
\email{masato.tamura.sf@hitachi.com},
\email{\{rahul.vishwakarma,ravigopal.vennelakanti\}@hal.hitachi.com}}
\maketitle

\begin{abstract}
This paper presents a novel framework for social group activity recognition. As an expanded task of group activity recognition, social group activity recognition requires recognizing multiple sub-group activities and identifying group members.
Most existing methods tackle both tasks by refining region features and then summarizing them into activity features. Such heuristic feature design renders the effectiveness of features susceptible to incomplete person localization and disregards the importance of scene contexts. Furthermore, region features are sub-optimal to identify group members
because the features may be dominated by those of people in the regions and have different semantics.
To overcome these drawbacks, we propose to leverage attention modules in transformers to generate effective social group features. Our method is designed in such a way that the attention modules identify and then aggregate features relevant to social group activities, generating an effective feature for each social group. Group member information is embedded into the features and thus accessed by feed-forward networks. The outputs of feed-forward networks represent groups so concisely that group members can be identified with simple Hungarian matching between groups and individuals.
Experimental results show that our method outperforms state-of-the-art methods on the Volleyball and Collective Activity datasets.

\end{abstract}

\section{Introduction}

\begin{figure}	
 \centering
 \begin{subfigure}[t]{1.\linewidth}
  \centering
  \includegraphics[width=1.0\linewidth]{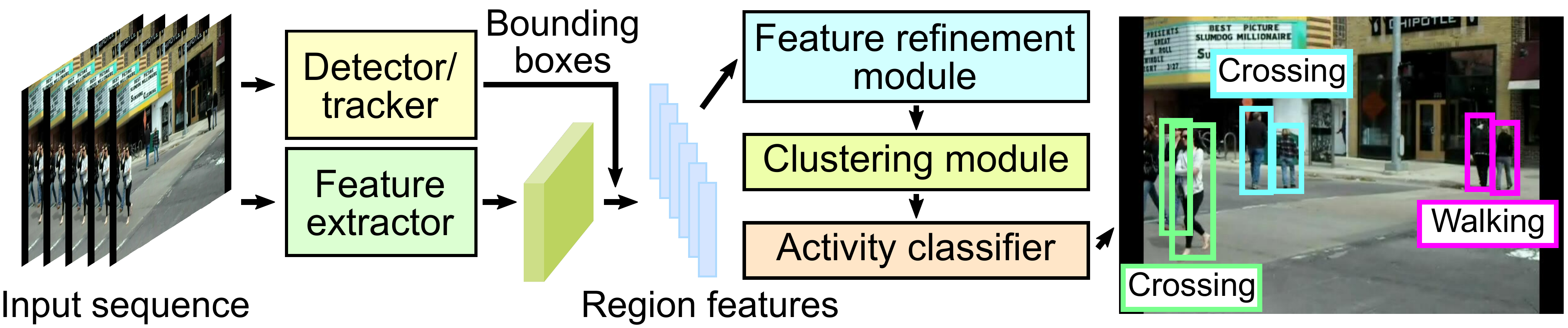}
  \caption{Conventional method.}\label{fig:conventional}
 \end{subfigure} \\
 \begin{subfigure}[t]{1.\linewidth}
  \centering
  \includegraphics[width=1.0\linewidth]{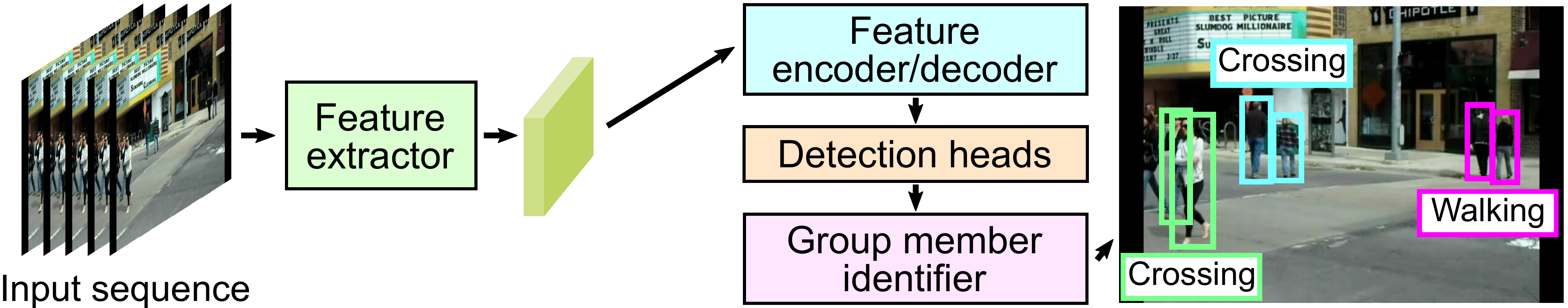}
  \caption{Proposed method.}\label{fig:proposed}
 \end{subfigure}
 \caption{Overviews of conventional and proposed social group activity recognition methods. The labels in the right image show predicted social group activities.}\label{fig:comp_overview}
 \vspace{-1.0ex}
\end{figure}

Social group activity recognition is a task of recognizing multiple sub-group activities and identifying group members in a scene. This task is derived from group activity recognition, which needs to recognize only one group activity in a scene. Both tasks have gained tremendous attention in recent years for potential applications such as sports video analysis, crowd behavior analysis, and social scene understanding~\cite{lan_nips2010,lan_cvpr2012,lan_tpami2012,amer_iccv2013,amer_eccv2014,amer_tpami2016,deng_cvpr2016,ibrahim_cvpr2016,bagautdinov_cvpr2017,shu_cvpr2017,wang_cvpr2017,li_iccv2017,kong_icassp2018,qi_eccv2018,ibrahim_eccv2018,wu_cvpr2019,azar_cvpr2019,gavrilyuk_cvpr2020,hu_cvpr2020,ehsanpour_eccv2020,pramono_eccv2020,li_iccv2021,yuan_iccv2021,Zhou2021COMPOSERCL}.
In the context of these tasks, the term ``action" denotes an atomic movement of a single person, and the term ``activity" refers to a more complex relation of movements performed by a group of people. Although our framework can recognize both actions and activities, we focus on group activities.

Most existing methods decompose the recognition process into two independent parts; person localization and activity recognition (See Fig.~\ref{fig:conventional})~\cite{deng_cvpr2016,ibrahim_cvpr2016,bagautdinov_cvpr2017,shu_cvpr2017,wang_cvpr2017,kong_icassp2018,qi_eccv2018,wu_cvpr2019,ehsanpour_eccv2020,gavrilyuk_cvpr2020,hu_cvpr2020,pramono_eccv2020,li_iccv2021,yuan_iccv2021,Zhou2021COMPOSERCL}.
Person localization identifies regions where people are observed in a scene with bounding boxes. These boxes are used to extract region features from feature maps. The region features are further refined to encode spatio-temporal relations with feature refinement modules such as recurrent neural networks (RNNs)~\cite{sepp_nc1998,cho_emnlp2014}, graph neural networks (GNNs)~\cite{kipf_iclr2017,velickovic_iclr2018}, and transformers~\cite{vaswani_nips2017}. The refined features are summarized for the purpose of activity recognition.

While these methods have demonstrated significant improvement, they have several drawbacks attributed to the heuristic nature of feature design.
Since region features are extracted from bounding box regions in feature maps, the effectiveness of the features is affected by the localization performance. Most existing methods ignore this effect and evaluate their performances with region features of ground truth boxes. However, several works~\cite{bagautdinov_cvpr2017,qi_eccv2018,wu_cvpr2019,ehsanpour_eccv2020} show that the recognition performance is slightly degraded when using predicted boxes instead of ground truth boxes.
Moreover, substantial scene contexts are discarded by using region features because they are typically dominated by features of the people in the boxes. Scene contexts such as object positions and background situations are sometimes crucial to recognize group activities. For instance, positions of sports balls are informative to recognize group activities in sports games. These features should be leveraged to enhance recognition performance.

Another challenge specific to social group activity recognition is that utilizing region features is sub-optimal to identify group members. Ehsanpour~\etal~\cite{ehsanpour_eccv2020} use region features as node features of graph attention networks (GATs)~\cite{velickovic_iclr2018} and train them to output adjacency matrices that have low probabilities for people in different groups and high probabilities for those in the same groups. During inference, spectral clustering~\cite{ng_nips2002} is applied to the adjacency matrices to divide people into groups. Because adjacency matrices reflect semantic similarities of node features, this method may not work if region features of people in the same group have different semantics such as doing different actions.

To address these challenges, we propose a novel social group activity recognition method that can be applied to both social group activity recognition and group activity recognition. We leverage a transformer-based object detection framework~\cite{carion_eccv2020,zhu_iclr2021} to obviate the need for the heuristic feature design in existing methods (See Fig.~\ref{fig:proposed}).
Attention modules in transformers play crucial roles in our method. We design our method in such a way that the attention modules identify and then aggregate features relevant to social group activities, generating an effective feature for each social group. Because activity and group member information is embedded into the generated features, the information can be accessed by feed-forward networks (FFNs) in the detection heads. The outputs of the detection heads are designed so concisely that group member identification can be performed with simple Hungarian matching between groups and individuals. This identification method differs from Ehsanpour~\etal's method~\cite{ehsanpour_eccv2020} in that their method relies on individuals' features to divide people into groups, while our method generates features that are embedded with clues for grouping people, enabling effective group identification.

To summarize, our contributions are three-fold:
\begin{itemize}
\item We propose a novel social group activity recognition method that leverages the attention modules in transformers to generate effective social group features. The group member information extracted from the features is designed to be concise and can be used to identify group members with a simple matching process.
\item Our method achieves better or competitive performance to state-of-the-art methods on both group activity recognition and social group activity recognition in two challenging benchmarks.
\item We perform comprehensive analyses to reveal how our method works with activities under various conditions.
\end{itemize}

\section{Related Works}

\subsection{Group Activity Recognition}
Deep-neural-network-based methods have become dominant in group activity recognition due to the learning capability of the networks.
Ibrahim~\etal~\cite{ibrahim_cvpr2016} proposed an RNN-based method that uses convolutional neural networks to extract features of person bounding box regions and long short-term memories to refine region features. This architecture captures the temporal dynamics of each person between frames and spatial dynamics of people in a scene. After their work, several RNN-based methods were proposed~\cite{shu_cvpr2017,bagautdinov_cvpr2017,wang_cvpr2017,kong_icassp2018,qi_eccv2018}.

GNNs are also utilized to model the spatio-temporal context and relationships of people in a scene.
Wu~\etal~\cite{wu_cvpr2019} used graph convolutional networks (GCNs)~\cite{kipf_iclr2017} to capture spatio-temporal relations of people's appearances and positions between frames.
Ehsanpour~\etal~\cite{ehsanpour_eccv2020} adopted GATs~\cite{velickovic_iclr2018} to learn underlying interactions and divide people into social groups with adjacency matrices.
Hu~\etal~\cite{hu_cvpr2020} utilized both RNNs and GNNs with reinforcement learning to refine features.
Yuan~\etal~\cite{yuan_iccv2021} used person-specific dynamic graphs that dynamically change connections of GNNs for each node.

With the rapid application of transformers~\cite{vaswani_nips2017} to vision problems, several works introduced transformers into group activity recognition. Gavrilyuk~\etal~\cite{gavrilyuk_cvpr2020} used transformer encoders to refine region features. Li~\etal~\cite{li_iccv2021} proposed spatial-temporal transformers that can encode spatio-temporal dependence and decode the group activity information. Zhou~\etal~\cite{Zhou2021COMPOSERCL} proposed multi-scale spatio-temporal stacked transformers for compositional understanding and relational reasoning in group activities.

Our method differs from existing methods in that they rely on region features, while our method generates social group features with the attention modules in transformers, resulting in improving the performance.

\subsection{Detection Transformer}
Carion~\etal~\cite{carion_eccv2020} proposed a transformer-based object detector called DETR, which regards object detection as a set prediction and achieves end-to-end object detection. One significant difference between conventional object detectors and DETR is that conventional ones need heuristic detection points whose features are used to predict object classes and bounding boxes, while DETR obviates such heuristic components by letting queries in transformer decoders aggregate features for their target objects with the attention mechanisms. DETR shows competitive performance compared with conventional state-of-the-art detectors even without such heuristic components.

To further improve the performance of DETR, several methods have been proposed~\cite{zhu_iclr2021,dai_iccv2021,sun_iccv2021}.
Zhu~\etal~\cite{zhu_iclr2021} proposed Deformable DETR that replaces standard transformers with deformable ones. Deformable attention modules in the transformers combine a sparse sampling of deformable convolution~\cite{dai_iccv2017} and dynamic weighting of standard attention modules, which significantly reduces the computational complexity of the attention weight calculation. This reduction allows Deformable DETR to use multi-scale feature maps from backbone networks.
To leverage non-heuristic designs and multi-scale feature maps, we use deformable transformers to generate social group features.

\section{Proposed Method}
We leverage a deformable-transformer-based object detection framework~\cite{zhu_iclr2021} to recognize multiple group activities and identify group members without the heuristic feature design.
We first explain the overall architecture in Sec.~\ref{subsec:overall} and show how we set up the framework for social group activity recognition.
In Sec.~\ref{subsec:loss_calc}, we describe the loss function used in training.
Finally, we explain the group member identification performed during inference in Sec.~\ref{subsec:grid}. Due to the limited space, we omit the details of deformable transformers and encourage readers to refer to the paper of Deformable DETR~\cite{zhu_iclr2021} for more details.

\subsection{Overall Architecture}\label{subsec:overall}

\begin{figure}[t]
 \centering
 \includegraphics[width=1.0\linewidth]{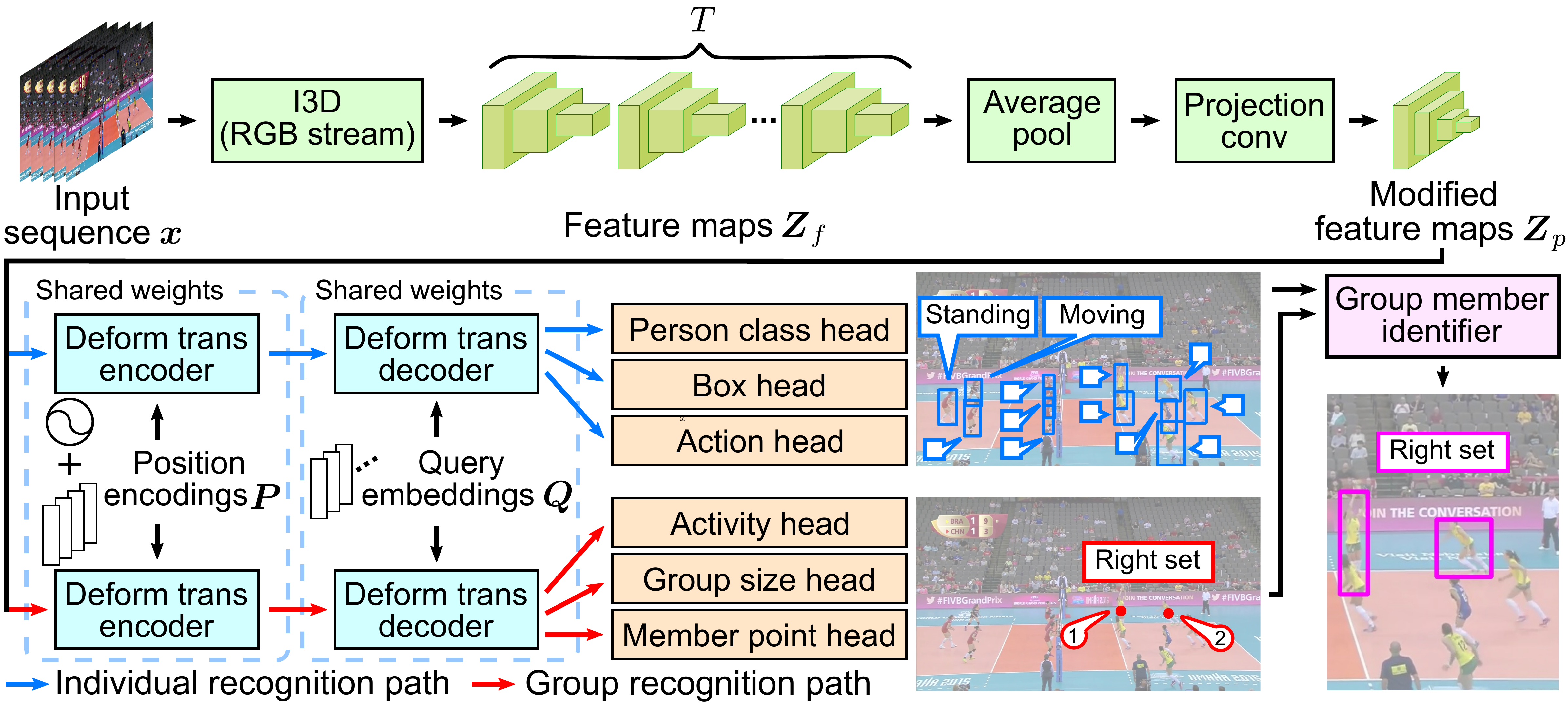}
 \caption{Overall architecture of the proposed method.}\label{fig:overview}
 \vspace{-1.0ex}
\end{figure}

Figure~\ref{fig:overview} shows the overall architecture of the proposed method.
Given a frame sequence $\bm{x} \in \mathbb{R}^{3 \times T \times H \times W}$, a feature extractor extracts a set of multi-scale feature maps $\bm{Z}_{f} = \{\bm{z}_{i}^{\left(f\right)} \mid \bm{z}_{i}^{\left(f\right)} \in \mathbb{R}^{D_{i} \times T \times H_{i}^{\prime} \times W_{i}^{\prime}}\}_{i=1}^{L_{f}}$, where $T$ is the length of the sequence, $H$ and $W$ are the height and width of the frame, $H_{i}^{\prime}$ and $W_{i}^{\prime}$ are those of the output feature maps, $D_{i}$ is the number of channels, and $L_{f}$ is the number of scales. We adopt the inflated 3D (I3D) network~\cite{carreira_cvpr2017} as a feature extractor to embed local spatio-temporal context into feature maps. Note that we use only the RGB stream of I3D because group members are identified by their positions, which cannot be predicted with the optical flow stream. To reduce the computational costs of transformers, each feature map $\bm{z}_{i}^{\left(f\right)}$ is mean-pooled over the temporal dimension and input to a projection convolution layer that reduces the channel dimension from $D_{i}$ to $D_{p}$. One additional projection convolution layer with a kernel size of $3 \times 3$ and stride of $2 \times 2$ is applied to the smallest feature map to further add the scale.

Features in the modified feature maps are refined and aggregated with deformable transformers.
Given a set of the modified multi-scale feature maps $\bm{Z}_{p} = \{\bm{z}_{i}^{\left(p\right)} \mid \bm{z}_{i}^{\left(p\right)} \in \mathbb{R}^{D_{p} \times H_{i}^{\prime} \times W_{i}^{\prime}}\}_{i=1}^{L_{f} + 1}$, a set of refined feature maps $\bm{Z}_{e} = \{\bm{z}_{i}^{\left(e\right)} \mid \bm{z}_{i}^{\left(e\right)} \in \mathbb{R}^{D_{p} \times H_{i}^{\prime} \times W_{i}^{\prime}}\}_{i=1}^{L_{f} + 1}$ is obtained as $\bm{Z}_{e} = f_{enc}\left(\bm{Z}_{p}, \bm{P}\right)$, where $f_{enc}\left(\cdot, \cdot\right)$ is stacked deformable transformer encoder layers and $\bm{P} = \{\bm{p}_{i} \mid \bm{p}_{i} \in \mathbb{R}^{D_{p} \times H_{i}^{\prime} \times W_{i}^{\prime}}\}_{i=1}^{L_{f} + 1}$ is a set of multi-scale position encodings~\cite{zhu_iclr2021}, which supplement the attention modules with position and scale information to identify where each feature lies in the feature maps. The encoder helps features to acquire rich social group context by exchanging information in a feature map and between multi-scale feature maps.
These enriched feature maps are fed into the deformable transformer decoder to aggregate features. Given a set of refined feature maps $\bm{Z}_{e}$ and learnable query embeddings $\bm{Q} = \{\bm{q}_{i} \mid \bm{q}_{i} \in \mathbb{R}^{2D_{p}}\}_{i=1}^{N_{q}}$, a set of feature embeddings $\bm{H} = \{\bm{h}_{i} \mid \bm{h}_{i} \in \mathbb{R}^{D_{p}}\}_{i=1}^{N_{q}}$ is obtained as $\bm{H} = f_{dec}\left(\bm{Z_{e}}, \bm{Q}\right)$, where $N_{q}$ is the number of query embeddings and $f_{dec}\left(\cdot, \cdot\right)$ is stacked deformable transformer decoder layers. Each decoder layer predicts locations that contain features relevant to input embeddings and aggregates the features from the locations with the dynamic weighting. We design queries in such a way that one query captures at most one social group. This design enables each query to aggregate features of its target social group from the refined feature maps.

The feature embeddings are transformed into prediction results with detection heads. Here we denote the localization results in normalized image coordinates.
Social group activities are recognized by predicting activities and identifying group members. The identification is performed with a group size head and group member point head. The size head predicts the number of people in a target social group, and the point head indicates group members by localizing the centers of group members' bounding boxes. This design enables our method to identify group members with simple point matching during inference as described in Sec.~\ref{subsec:grid}. The predictions of activity class probabilities $\{\bm{\hat{v}}_{i} \mid \bm{\hat{v}}_{i} \in [0, 1]^{N_{v}}\}_{i=1}^{N_q}$, group sizes $\{\hat{s}_{i} \mid \hat{s}_{i} \in [0, 1]\}_{i=1}^{N_q}$, and sequences of group member points $\{\bm{\hat{U}}_{i}\}_{i=1}^{N_q}$ are obtained as $\bm{\hat{v}}_{i} = f_{v}\left(\bm{h}_{i}\right)$, $\hat{s}_{i} = f_{s}\left(\bm{h}_{i}\right)$, and $\bm{\hat{U}}_{i} = f_{u}\left(\bm{h}_{i}, \bm{r}_{i}\right)$, where $N_{v}$ is the number of activity classes, $\bm{\hat{U}}_{i} = (\bm{\hat{u}}^{(i)}_{j} \mid \bm{\hat{u}}^{(i)}_{j} \in [0, 1]^{2})^{M}_{j = 1}$ is a sequence of points that indicate centers of group members' bounding boxes, $M$ is a hyper-parameter that defines the maximum group size, $f_{v}\left(\cdot\right)$, $f_{s}\left(\cdot\right)$, and $f_{u}\left(\cdot, \cdot\right)$ are the detection heads for each prediction, and $\bm{r}_{i} \in [0, 1]^2$ is a reference point, which is used in the same way as the localization in Deformable DETR~\cite{zhu_iclr2021}. The predicted group sizes are values normalized with $M$. All the detection heads are composed of FFNs with subsequent sigmoid functions. We describe the details of the detection heads in the supplementary material.

Individual recognition can be performed by replacing the group recognition heads with individual recognition heads. We empirically find that using different parameters of deformable transformers for individual recognition and social group recognition does not show performance improvement and thus use shared parameters to reduce the computational costs. The details of the individual recognition heads are described in the supplementary material. 

\subsection{Loss Calculation}\label{subsec:loss_calc}

We view social group activity recognition as a direct set prediction problem and match predictions and ground truths with the Hungarian algorithm~\cite{kuhn_naval1955} during training following the training procedure of DETR~\cite{carion_eccv2020}.
The optimal assignment is determined by calculating the matching cost with the predicted activity class probabilities, group sizes, and group member points. Given a ground truth set of social group activity recognition, the set is first padded with $\phi^{(gr)}$ (no activity) to change the set size to $N_{q}$. With the padded ground truth set, the matching cost of $i$-th element in the ground truth set and $j$-th element in the prediction set is calculated as follows:
\begin{align}
  \mathcal{H}^{(gr)}_{i, j} ={} & \mathbbm{1}_{\{i \not\in \bm{\Phi}^{(gr)}\}}\left[\eta_{v} \mathcal{H}^{(v)}_{i, j} + \eta_{s} \mathcal{H}^{(s)}_{i, j} + \eta_{u} \mathcal{H}^{(u)}_{i, j}\right], \\
  \mathcal{H}^{(v)}_{i, j} ={} & -\frac{\bm{v}^{T}_{i}\bm{\hat{v}}_{j} + \left(\bm{1} - \bm{v}_{i}\right)^{T}\left(\bm{1} - \bm{\hat{v}}_{j}\right)}{N_{v}}, \\
  \mathcal{H}^{(s)}_{i, j} ={} & \left|s_{i} - \hat{s}_{j}\right|, \\
  \mathcal{H}^{(u)}_{i, j} ={} & \frac{\sum_{k = 1}^{S_{i}} \left\|\bm{u}^{(i)}_{k} - \bm{\hat{u}}^{(j)}_{k}\right\|_{1}}{S_{i}},
\end{align}
where $\bm{\Phi}^{(gr)}$ is a set of ground-truth indices that correspond to $\phi^{(gr)}$, $\bm{v}_{i} \in \{0, 1\}^{N_{v}}$ is a ground truth activity label, $s_{i} \in [0, 1]$ is a ground truth group size normalized with $M$, $S_{i}$ is an unnormalized ground truth group size, $\bm{u}^{(i)}_{k} \in [0, 1]^{2}$ is a ground truth group member point normalized with the image size, and $\eta_{\{v, s, u\}}$ are hyper-parameters. Group member points in the sequence $\bm{U}_{i} = (\bm{u}^{(i)}_{k})^{S_{i}}_{k = 1}$ are sorted in ascending order along $X$ coordinates as seen from the image of the group recognition result in Fig.~\ref{fig:overview}. We use this arrangement because group members are typically seen side by side at the same vertical positions in an image, and the order of group member points is clear from their positions, which makes the prediction easy. We evaluate the performances with other arrangements and compare the results in Sec.~\ref{subsec:order}. Using Hungarian algorithm, the optimal assignment is calculated as $\hat{\omega}^{(gr)} = \argmin_{\omega \in \bm{\Omega}_{N_q}}{\sum_{i=1}^{N_q}{\mathcal{H}^{(gr)}_{i,\omega(i)}}}$, where $\bm{\Omega}_{N_q}$ is the set of all possible permutations of $N_q$ elements.

The training loss for social group activity recognition $\mathcal{L}_{gr}$ is calculated between matched ground truths and predictions as follows:
\begin{align}
  \mathcal{L}_{v} ={} & \frac{1}{|\bar{\bm{\Phi}}^{(gr)}|} \sum_{i=1}^{N_{q}}\left[
          \mathbbm{1}_{\{i \not\in \bm{\Phi}^{(gr)}\}}l_{f}\left(\bm{v}_{i}, \bm{\hat{v}}_{\hat{\omega}^{(gr)}\left(i\right)}\right) + \mathbbm{1}_{\{i \in \bm{\Phi}^{(gr)}\}}l_{f}\left(\bm{0}, \bm{\hat{v}}_{\hat{\omega}^{(gr)}\left(i\right)}\right)\right], \\
  \mathcal{L}_{s} ={} & \frac{1}{|\bar{\bm{\Phi}}^{(gr)}|} \sum_{i=1}^{N_{q}} \mathbbm{1}_{\{i \not\in \bm{\Phi}^{(gr)}\}}\left|s_i - \hat{s}_{\hat{\omega}^{(gr)}\left(i\right)}\right|, \\
  \mathcal{L}_{u} ={} & \frac{1}{|\bar{\bm{\Phi}}^{(gr)}|} \sum_{i=1}^{N_{q}} \sum_{j=1}^{S_{i}} \mathbbm{1}_{\{i \not\in \bm{\Phi}^{(gr)}\}} \left\|\bm{u}^{(i)}_j - \bm{\hat{u}}^{(\hat{\omega}^{(gr)}\left(i\right))}_{j}\right\|_{1},
\end{align}
where $\lambda_{\{v, s, u\}}$ are hyper-parameters and $l_{f}\left(\cdot, \cdot\right)$ is the element-wise focal loss function~\cite{lin_iccv2017} whose hyper-parameters are described in~\cite{zhou_arxiv2019}.

Individual recognition is jointly learned by matching ground truths and predictions of person class probabilities, bounding boxes, and action class probabilities and calculating the losses between matched ground truths and predictions. The matching and loss calculations are performed by slightly modifying the original matching costs and losses of Deformable DETR~\cite{zhu_iclr2021}. We describe the details of these matching and loss calculations in the supplementary material.

\subsection{Group Member Identification}\label{subsec:grid}
The outputs of the detection heads represent groups in group sizes and group member points that indicate centers of group members' bounding boxes. These values have to be transformed into values that indicate individuals. We transform the predicted values into indices that refer to the elements in the individual prediction set with the following simple process during inference.
To match the group member points and individual predictions, the Hungarian algorithm~\cite{kuhn_naval1955} is used instead of just calculating the closest center of a bounding box for each group member point. Hungarian algorithm can prevent multiple group member points from matching the same individuals and thus slightly improve the performance. The matching cost between $i$-th group member point of $k$-th social group prediction and $j$-th individual prediction is calculated as follows:
\begin{equation}
  \mathcal{H}^{(gm, k)}_{i, j} = \frac{\left\|\bm{\hat{u}}^{(k)}_{i} - f_{cent}\left(\bm{\hat{b}}_{j}\right)\right\|_{2}}{\hat{c}_{j}},
\end{equation}
where $\bm{\hat{b}}_{j} \in [0, 1]^{4}$ is a predicted bounding box of an individual, $\hat{c}_{j} \in [0, 1]$ is a detection score of the individual, and $f_{cent}\left(\cdot\right)$ is a function that calculates the center of a bounding box. By applying the Hungarian algorithm to this matching cost, the optimal assignment is calculated as $\hat{\omega}^{(gm, k)} = \argmin_{\omega \in \bm{\Omega}_{N_q}}{\sum_{i=1}^{\left\lfloor M \times \hat{s}_{k}\right\rceil}{\mathcal{H}^{(gm, k)}_{i,\omega(i)}}}$, where $\left\lfloor \cdot\right\rceil$ rounds an input value to the nearest integer. Finally, the index set of individuals for $k$-th social group prediction is obtained as $\bm{G}_{k} = \{\hat{\omega}^{(gm, k)}\left(i\right)\}^{\left\lfloor M \times \hat{s}_{k}\right\rceil}_{i=1}$.

\section{Experiments}

\subsection{Datasets and Evaluation Metrics}\label{subsec:dataset}
We evaluate the performance of our method on two publicly available benchmark datasets: Volleyball dataset~\cite{ibrahim_cvpr2016} and Collective Activity dataset~\cite{choi_iccvw2009}.
The Volleyball dataset contains 4,830 videos of 55 volleyball matches, which are split into 3,493 training videos and 1,337 test videos. The center frame of each video is annotated with bounding boxes, actions, and one group activity. The number of action and activity classes are 9 and 8, respectively. Because the original annotations do not contain group member information, we use an  extra annotation set provided by Sendo and Ukita~\cite{sendo_mva2019}. We combine the original annotations with the group annotations in the extra set  and use them for our experiments. Note that annotations other than the group annotations in the extra set are not used for a fair comparison.
The Collective Activity dataset contains 44 videos of life scenes, which are split into 32 training videos and 12 test videos. The videos are annotated every ten frames with bounding boxes and actions. The group activity is defined as the action with the largest number in the scenes. The number of action classes is 6. Because the original annotations do not have group member information, Ehsanpour~\etal~\cite{ehsanpour_eccv2020} annotated group labels. We use their annotations for our experiments.

We divide the evaluation into two parts: group activity recognition and social group activity recognition.
In the evaluation of group activity recognition, we follow the detection-based settings~\cite{bagautdinov_cvpr2017,qi_eccv2018,wu_cvpr2019,ehsanpour_eccv2020} and use classification accuracy as an evaluation metric. Because our method is designed to predict multiple group activities, we need to select one from them for group activity recognition. We choose the predicted activity of the highest probability and compare it with the ground truth activity.
In the evaluation of social group activity recognition, different metrics are used for each dataset because each scene in the Volleyball dataset contains only one social group activity, while that in the Collective Activity dataset contains multiple social group activities. For the Volleyball dataset, group identification accuracy is used as an evaluation metric. One group prediction is first selected in the same way as group activity recognition, and then the predicted bounding boxes of the group members are compared with the ground truth boxes. The selected prediction results are correct if the predicted activity is correct and the predicted boxes have IoUs larger than 0.5 with the corresponding ground truth boxes. For the Collective Activity dataset, mAP is used as an evaluation metric. Prediction results are judged as true positives if the predicted activities are correct, and all the predicted boxes of the group members have IoUs larger than 0.5 with the corresponding ground truth boxes.

\subsection{Implementation Details}

We use the RGB stream of I3D~\cite{carreira_cvpr2017} as a backbone feature extractor and input features from \textit{Mixed\_3c}, \textit{Mixed\_4f}, and \textit{Mixed\_5c} layers into the deformable transformers. The hyper-parameters of the deformable transformers are set in accordance with the setting of Deformable DETR~\cite{zhu_iclr2021}, where $L_{f} = 3$, $D_{p} = 256$, and $N_{q} = 300$. We initialize I3D with the parameters trained on the Kinetics dataset~\cite{kay_arxiv2017} and deformable transformers with the parameters trained on the COCO dataset~\cite{lin_eccv2014}. We use the AdamW~\cite{loshchiloy_iclr2019} optimizer with the batch size of 16, the initial learning rate of $10^{-4}$, and the weight decay of $10^{-4}$. Training epochs are set to 120, and the learning rate is decayed after 100 epochs. We set the length of the sequence $T$ to 9. Ground truth labels of the center frame are used to calculate the losses. To augment the training data, we randomly shift frames in the temporal direction and use bounding boxes from visual trackers as ground truth boxes when a non-annotated frame is at the center. We also augment the training data by random horizontal flipping, scaling, and cropping. Following the DETR's training~\cite{carion_eccv2020}, auxiliary losses are used to boost the performance. The maximum group size $M$ is set to 12. The hyper-parameters are set as $\eta_{v} = \lambda_{v} = 2$, $\eta_{s} = \lambda_{s} = 1$, and $\eta_{u} = \lambda_{u} = 5$.

While evaluating performances with the Collective Activity dataset, some specific settings are used. For the evaluation of group activity recognition, training epochs are set to 10, and the learning rate is decayed after 5 epochs because the losses converge in a few epochs due to the limited diversity of the scenes in the dataset. For the evaluation of social group activity recognition, the length of the sequence $T$ is set to 17 following the setting of Ehsanpour~\etal~\cite{ehsanpour_eccv2020}.

\subsection{Group Activity Recognition}

\subsubsection{Comparison against State-of-the-Art.}
\begin{table}[t]
 \caption{Comparison against state-of-the-art methods on group activity recognition. The values with and without the brackets demonstrate the performances in the ground-truth-based and detection-based settings, respectively. The performances of individual action recognition are shown for future reference.}
 \label{table:comp_indgroupact}
 \centering
 \setlength{\tabcolsep}{3.2pt}
 \begin{threeparttable}
  \begin{tabular}{@{}lcccccccc@{}}
   \toprule
   & \multicolumn{4}{c}{Volleyball} & \multicolumn{4}{c}{Collective Activity} \\
   \cmidrule(lr){2-5}\cmidrule(lr){6-9}
   Method & \multicolumn{2}{c}{Activity} & \multicolumn{2}{c}{Action} & \multicolumn{2}{c}{Activity} & \multicolumn{2}{c}{Action} \\
   \midrule
   SSU~\cite{bagautdinov_cvpr2017} & 86.2 & (90.6) & -- & (81.8) & -- & (\hspace{0.6em}--\hspace{0.6em}) & -- & (\hspace{0.6em}--\hspace{0.6em}) \\
   stagNet~\cite{qi_eccv2018} & 87.6 & (89.3) & -- & (\hspace{0.6em}--\hspace{0.6em}) & 87.9 & (89.1) & -- & (\hspace{0.6em}--\hspace{0.6em}) \\
   ARG~\cite{wu_cvpr2019} & 91.5 & (92.5) & 39.8 & (83.0) & 86.1 & (88.1) & 49.6 & (77.3) \\
   CRM~\cite{azar_cvpr2019} & -- & (93.0) & -- & (\hspace{0.6em}--\hspace{0.6em}) & -- & (85.8) & -- & (\hspace{0.6em}--\hspace{0.6em}) \\
   PRL~\cite{hu_cvpr2020} & -- & (91.4) & -- & (\hspace{0.6em}--\hspace{0.6em}) & -- & (\hspace{0.6em}--\hspace{0.6em}) & -- & (\hspace{0.6em}--\hspace{0.6em}) \\
   Actor-Transformers~\cite{gavrilyuk_cvpr2020} & -- & (94.4) & -- & (85.9) & -- & (92.8) & -- & (\hspace{0.6em}--\hspace{0.6em}) \\
   Ehsanpour~\etal~\cite{ehsanpour_eccv2020} & 93.0 & (93.1) & 41.8 & (83.3) & 89.4 & (89.4) & 55.9 & (78.3) \\
   Pramono~\etal~\cite{pramono_eccv2020} & -- & (95.0) & -- & (83.1) & -- & (95.2) & -- & (\hspace{0.6em}--\hspace{0.6em}) \\
   DIN~\cite{yuan_iccv2021} & -- & (93.6) & -- & (\hspace{0.6em}--\hspace{0.6em}) & -- & (95.9) & -- & (\hspace{0.6em}--\hspace{0.6em}) \\
   GroupFormer~\cite{li_iccv2021} & 95.0\tnote{*} & (95.7) & -- & (85.6) & 85.2\tnote{*} & (87.5\tnote{\dag} /96.3) & -- & (\hspace{0.6em}--\hspace{0.6em}) \\
   \midrule
   Ours & \textbf{96.0} & (\hspace{0.6em}--\hspace{0.6em}) & \textbf{65.0} & (\hspace{0.6em}--\hspace{0.6em}) & \textbf{96.5} & (\hspace{0.6em}--\hspace{0.6em}) & \textbf{64.9} & (\hspace{0.6em}--\hspace{0.6em}) \\
   \bottomrule
  \end{tabular}
  \begin{tablenotes}\footnotesize
   \item[*] We evaluated the performance with the publicly available source codes.
   \item[\dag] We evaluated but were not able to reproduce the reported accuracy because the configuration file for the Collective Activity dataset is not publicly available.
  \end{tablenotes}
 \end{threeparttable}
 \vspace{-1.0em}
\end{table}

We compare our method against state-of-the-art methods on group activity recognition. Table~\ref{table:comp_indgroupact} shows the comparison results. The values without the brackets demonstrate the detection-based performances, while those inside the brackets indicate the performances with ground truth bounding boxes. We show the performances of individual action recognition for future reference. Several detection-based performances are not reported because existing works typically use ground-truth boxes for the evaluation. To compare the effectiveness with these methods, we evaluate GroupFormer~\cite{li_iccv2021}, which is the strongest baseline of group activity recognition, with predicted boxes of Deformable DETR~\cite{zhu_iclr2021}. Note that Deformable DETR is fine-tuned on each dataset for a fair comparison, which demonstrates 90.8 and 90.2 mAP on the Volleyball and Collective Activity datasets, respectively.

As seen from the table, our method outperforms state-of-the-art methods in the detection-based setting. We confirm that GroupFormer shows the performance degradation as well as the previous methods~\cite{bagautdinov_cvpr2017,qi_eccv2018,wu_cvpr2019,ehsanpour_eccv2020} when predicted bounding boxes are used. These results indicate that the latest region-feature-based method still suffers from incomplete person localization and that our feature generation has advantages over these methods.
Even compared to the ground-truth-based performances, our method shows the best performance. It is worth noting that our method uses only RGB images as inputs, while GroupFormer utilizes optical flows and pose information in addition to RGB data. These results suggest that features generated by our method are more effective than region features and that it is not optimal to restrict regions of features to bounding boxes.

\subsubsection{Analysis on Group Annotations.}
\begin{table}[t]
 \caption{Analysis on the effect of the group annotations with the Volleyball dataset. The values with and without the brackets demonstrate the performances in the ground-truth-based and detection-based settings, respectively.}
 \label{table:analysis_group_annos}
 \centering
 \setlength{\tabcolsep}{6pt}
 \begin{threeparttable}
  \begin{tabular}{@{}l@{\hskip 6em}c@{\hskip 7em}rl@{}}
   \toprule
   Method & Annotation type & \multicolumn{2}{c}{Activity} \\
   \midrule
   \multirow{2}{*}{GroupFormer~\cite{li_iccv2021}} & Original & 95.0\tnote{*} & (95.7) \\
   & Group & 93.2\tnote{\ddag} & (96.1\tnote{*}\hspace{0.4em}) \\
   \midrule
   \multirow{2}{*}{Ours} & Original & 95.0 & (\hspace{0.6em}--\hspace{0.6em}) \\
   & Group & \textbf{96.0} & (\hspace{0.6em}--\hspace{0.6em}) \\
   \bottomrule
  \end{tabular}
  \begin{tablenotes}\footnotesize
    \item[*] We evaluated the performance with the publicly available source codes.
    \item[\ddag] We trained a group member detector and evaluated the performance with publicly available source codes.
  \end{tablenotes}
 \end{threeparttable}
 \vspace{-1.0em}
\end{table}

As described in Sec.~\ref{subsec:dataset}, we use the additional group annotations to fully leverage our social group activity recognition capability. We analyze the effect of the group annotations on group activity recognition by investigating the performances of both GroupFormer~\cite{li_iccv2021} and our method with and without the group annotations. Note that hereinafter we use the Volleyball dataset for analyses because the diversity of the scenes in the Collective Activity dataset is limited. To evaluate GroupFormer with the group annotations in the detection-based setting, we trained Deformable DETR~\cite{zhu_iclr2021} with bounding boxes of only group members, which is intended to detect only people involved in activities. The detector shows the performance of 87.1 mAP.
Among all the results, GroupFormer with the group annotations in the ground-truth-based setting demonstrates the best performance. However, the performance is substantially degraded when the predicted boxes are used. This is probably because group member detection underperforms and degrades the recognition performance. As our method does not rely on bounding boxes to predict group activities, the performance does not degrade even if group members cannot be identified correctly. Accordingly, our method demonstrates the best performance in the detection-based setting.

\subsection{Social Group Activity Recognition}

\subsubsection{Comparison against State-of-the-Art.}
\begin{table}[t]
 \caption{Comparison against state-of-the-art social group activity recognition methods with the Volleyball dataset.}
 \label{table:comp_socialgroup_volley}
 \centering
 \setlength{\tabcolsep}{2.2pt}
 \begin{threeparttable}
  \begin{tabular}{@{}lccccccccc@{}}
   \toprule
   & & \multicolumn{4}{c}{Right} & \multicolumn{4}{c}{Left} \\
   \cmidrule(lr){3-6}\cmidrule(lr){7-10}
   Method & Accuracy & Set & Spike & Pass & Winpoint & Set & Spike & Pass & Winpoint \\
   \midrule
   Ehsanpour~\etal~\cite{ehsanpour_eccv2020}\tnote{\S} & 44.5 & 17.2 & \textbf{74.0} & 49.0 & 29.9 & 19.7 & \textbf{79.6} & 25.0 & 28.4 \\
   GroupFormer~\cite{li_iccv2021}\tnote{\ddag} & 48.8 & 25.0 & 56.6 & 59.0 & \textbf{51.7} & 31.5 & 55.3 & 58.8 & 51.0 \\
   \midrule
   Ours & \textbf{60.6} & \textbf{35.9} & 68.2 & \textbf{81.9} & 50.6 & \textbf{50.6} & 53.6 & \textbf{74.3} & \textbf{56.9} \\
   \bottomrule
  \end{tabular}
  \begin{tablenotes}\footnotesize
    \item[\S] Because the source codes are not publicly available, we implemented their algorithm based on our best understanding and evaluated the performance.
    \item[\ddag] We trained a group member detector and evaluated the performance with publicly available source codes.
  \end{tablenotes}
 \end{threeparttable}
 \vspace{-1.0em}
\end{table}

\begin{table}[t]
 \caption{Comparison against a state-of-the-art social group activity recognition method with the Collective Activity dataset.}
 \label{table:comp_socialgroup_collective}
 \centering
 \setlength{\tabcolsep}{6pt}
 \begin{tabular}{@{}lcccccc@{}}
  \toprule
  Method & mAP & Crossing & Waiting & Queueing & Walking & Talking \\
  \midrule
  Ehsanpour~\etal~\cite{ehsanpour_eccv2020} & \textbf{51.3} & -- & -- & -- & -- & -- \\
  \midrule
  Ours & 46.0 & 49.2 & 64.5 & 54.1 & 55.6 & 6.56 \\
  \bottomrule
 \end{tabular}
 \vspace{-1.0em}
\end{table}

To demonstrate the effectiveness of our method on social group activity recognition, we compare our method against Ehsanpour~\etal's method~\cite{ehsanpour_eccv2020}, which is a state-of-the-art method that tackles social group activity recognition, and GroupFormer~\cite{li_iccv2021}, which is the strongest baseline on group activity recognition. Due to the unavailability of both Ehsanpour~\etal's source codes and their performance report on the Volleyball dataset, we implemented their algorithm based on our best understanding and evaluated the performance on the dataset. For the evaluation of GroupFormer, we trained Deformable DETR~\cite{zhu_iclr2021} in the same way as described in the group annotation analysis section for detecting group members. Because this group member detection cannot be applied to multiple social groups, we evaluate GroupFormer only on the Volleyball dataset.

Table~\ref{table:comp_socialgroup_volley} shows the results on the Volleyball dataset. As shown in the table, our method yields significant performance gains over the other methods, which demonstrates the improvement on group member identification as well as on activity recognition. Our method aggregates features that are embedded with clues for grouping people from feature maps. It is highly likely that this feature aggregation contributes to the high accuracy of identifying activities with different distributions of group members in an image. We qualitatively analyze how features are aggregated depending on the distribution of group members and discuss the analysis results towards the end in the qualitative analysis section.

The comparison results on the Collective Activity dataset are listed in Table~\ref{table:comp_socialgroup_collective}. As seen from the table, Ehsanpour\etal's method shows better performance than our method. We find that our method demonstrates relatively low performance on the activity ``Talking''. This low performance is probably attributed to the number of samples in training data. In the test data, \SI{86}{\percent} of samples with the activity ``Talking'' have the group sizes of four, while the training data has only 57 samples whose group sizes are four, which is \SI{0.8}{\percent} of the training data. As our method learns to predict group sizes, the number of samples in training data for each group size affects the performance. We analyze this effect in the subsequent section.

\subsubsection{Analysis on Group Sizes.}
\begin{table}[t]
 \caption{Analysis on group sizes with Volleyball dataset.}
 \label{table:analysis_gsize}
 \centering
 \setlength{\tabcolsep}{5pt}
 \begin{threeparttable}
  \begin{tabular}{@{}lcccccc@{}}
   \toprule
   & \multicolumn{6}{c}{Group size (Training data ratio)} \\
   Method & 1 (\SI{36}{\percent}) & 2 (\SI{21}{\percent}) & 3 (\SI{19}{\percent}) & 4 (\SI{6}{\percent}) & 5 (\SI{5}{\percent}) & 6 (\SI{12}{\percent}) \\
   \midrule
   Ehsanpour~\etal~\cite{ehsanpour_eccv2020}\tnote{\S} & 45.3 & \textbf{48.2} & \textbf{61.2} & 27.3 & 15.8 & 32.5 \\
   GroupFormer~\cite{li_iccv2021}\tnote{\ddag} & 57.3 & 29.6 & 58.4 & \textbf{28.4} & \textbf{44.7} & 54.4 \\
   \midrule
   Ours & \textbf{83.6} & 42.9 & 52.4 & 26.1 & 39.5 & \textbf{63.8} \\
   \bottomrule
  \end{tabular}
  \begin{tablenotes}\footnotesize
   \item[\S] Because the source codes are not publicly available, we implemented their algorithm based on our best understanding and evaluated the performance.
   \item[\ddag] We trained a group member detector and evaluated the performance with publicly available source codes.
  \end{tablenotes}
 \end{threeparttable}
 \vspace{-1.0em}
\end{table}

The group size prediction is one of the key factors to identify group members and thus affects social group activity recognition performance. To analyze this effect, we evaluate the performance on each group size and compare the results with Ehsanpour\etal's method~\cite{ehsanpour_eccv2020} and GroupFormer~\cite{li_iccv2021}.
Table~\ref{table:analysis_gsize} shows the results. As shown in the table, the performances of our method are moderately correlated to the training data ratios, while the other two methods do not show the correlation. This is the drawback of our method that relies on group size learning. However, our method shows the competitive performances on both small and large group sizes if there are a certain amount of training data. In contrast, each of the other two methods shows the competitive performances only on either large or small group sizes. These results imply that our method does not have the performance dependence on group sizes and thus can achieve high performance with large-scale training data.
 
\subsubsection{Analysis on Order of Group Member Points.}\label{subsec:order}
\begin{table}[t]
  \caption{Analysis on the order of member points with the Volleyball dataset.}
  \label{table:analysis_order}
  \centering
  \setlength{\tabcolsep}{8pt}
  \begin{tabular}{@{}lcc@{}}
   \toprule
   Order of the group member points & Probability of changes in order & Accuracy \\
   \midrule
   Ascending order in X coordinates & \SI{7.4}{\percent} & \textbf{60.6} \\
   Ascending order in Y coordinates & \SI{13}{\percent} & 55.5 \\
   \bottomrule
  \end{tabular}
  \vspace{-1.0em}
\end{table}

As described in Sec.~\ref{subsec:loss_calc}, group member points in a ground truth point sequence are sorted in ascending order along $X$ coordinates. To confirm the effectiveness of this arrangement, we compare the performances with two arrangements. Table~\ref{table:analysis_order} shows the comparison results. As shown in the table, our method demonstrates better performance when group member points are sorted in ascending order along $X$ coordinates than in ascending order along $Y$ coordinates. The probabilities in the table indicate the ratio of the changes of the point order when small perturbations are added to ground-truth bounding box positions. The higher probability implies that the order of group member points changes more frequently when group members move. These results suggest that the order changes more frequently when group member points are sorted in ascending order along $Y$ coordinates and that the order is difficult to predict with slight differences of box positions.
 
\subsubsection{Qualitative Analysis.}

\begin{figure}[t]
 \centering
 \includegraphics[width=1.0\linewidth]{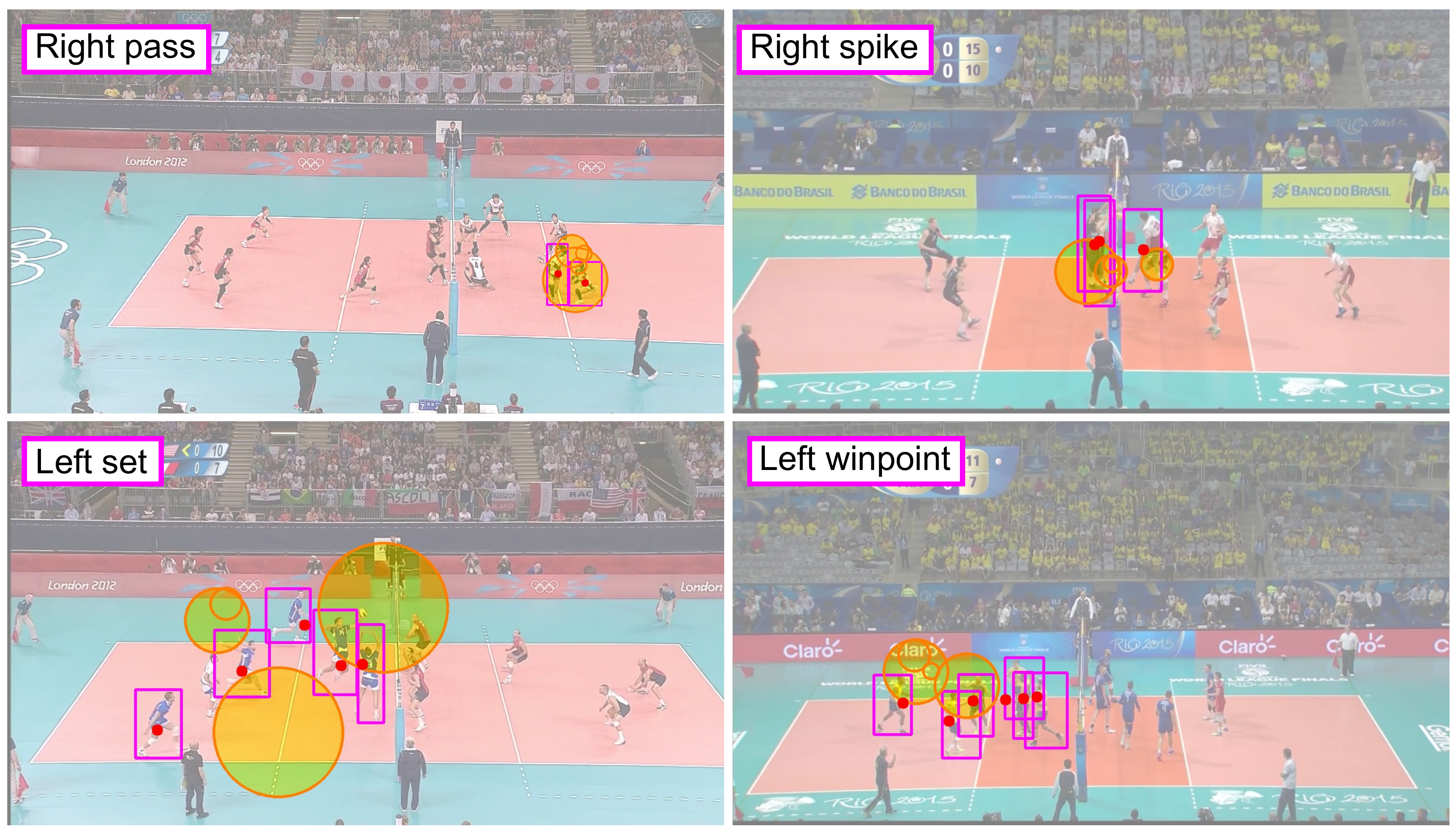}
 \caption{Visualization of the attention locations in the deformable transformer decoder. We show the locations of the top four attention weights. The large circles mean that the locations are in the low-resolution feature maps.}\label{fig:qualitative}
 \vspace{-1.0ex}
\end{figure}

The deformable attention modules are the critical components to aggregate features relevant to social group activity recognition and generate social group features. To analyze how the attention modules aggregate features for various social group activities, we visualize the attention locations of the transformer decoder in Fig.~\ref{fig:qualitative}. We show locations with the top four attention weights in the last layer of the decoder. The purple bounding boxes show the group members, the red circles show the predicted group member points, and the yellow circles show the attention locations. The small and large yellow circles mean that the locations are in the high and low-resolution feature maps, respectively, showing a rough range of image areas affecting the generated features. The figure shows that features are typically aggregated from low-resolution feature maps if group members are located in broad areas, and vice versa. These results indicate that the attention modules can effectively aggregate features depending on the distribution of group members and contribute to improving the performance of social group activity recognition.

\section{Conclusions}
We propose a novel social group activity recognition method that leverages deformable transformers to generate effective social group features. 
This feature generation obviates the need for region features and hence makes the effectiveness of the social group features person-localization-agnostic.
Furthermore, the group member information extracted from the features is represented so concisely that our method can identify group members with simple Hungarian matching, resulting in high-performance social group activity recognition. 
We perform extensive experiments and show significant improvement over existing methods.

\section*{Acknowledgement}
Computational resource of AI Bridging Cloud Infrastructure (ABCI) provided by National Institute of Advanced Industrial Science and Technology (AIST) was used.

\bibliographystyle{splncs04}
\bibliography{main}

\renewcommand{\thesection}{\Alph{section}}
\setcounter{section}{0}

\clearpage

\centerline{\Large \bfseries\boldmath Supplementary Material}

\section{Individual Recognition}

In our method, individuals are recognized by simply adding an action classification head to the detection heads in Deformable DETR~\cite{zhu_iclr2021}. Given a set of feature embedding $\bm{H} = \{\bm{h}_{i} \mid \bm{h}_{i} \in \mathbb{R}^{D_{p}}\}_{i=1}^{N_{q}}$ from the deformable transformer decoder, the predictions of person class probabilities $\{\hat{c}_{i} \mid \hat{c}_{i} \in [0, 1]\}_{i=1}^{N_q}$, bounding boxes $\{\bm{\hat{b}}_{i} \mid \bm{\hat{b}}_{i} \in [0, 1]^4\}_{i=1}^{N_q}$, and action class probabilities $\{\bm{\hat{a}}_{i} \mid \bm{\hat{a}}_{i} \in [0, 1]^{N_{a}}\}_{i=1}^{N_q}$ are obtained as $\hat{c}_{i} = f_{c}\left(\bm{h}_{i}\right)$, $\bm{\hat{b}}_{i} = f_{b}\left(\bm{h}_{i}, \bm{r}_{i}\right)$, and $\bm{\hat{a}}_{i} = f_{a}\left(\bm{h}_{i}\right)$, where $N_{q}$ is the number of query embeddings, $N_{a}$ is the number of action classes, $f_{c}\left(\cdot\right)$, $f_{b}\left(\cdot, \cdot\right)$, and $f_{a}\left(\cdot\right)$ are the detection heads for the predictions, and $\bm{r}_{i} \in [0, 1]^2$ is a reference point, which is used in the same way as the localization in Deformable DETR. Note that the localization results are denoted in the normalized image coordinates. 

We view individual recognition as a direct set prediction problem and match predictions and ground truths with the Hungarian algorithm~\cite{kuhn_naval1955} during training. The optimal assignment of ground truths and predictions is determined by calculating the matching cost with the predicted person class probabilities, bounding boxes, and action class probabilities. Given a ground truth set of individual recognition, the set is first padded with $\phi^{(id)}$ (no person) to change the size of the set to $N_{q}$. Using the padded ground truth set, the matching cost of $i$-th element in the ground truth set and $j$-th element in the prediction set for individual recognition is calculated as follows:
\begin{align}
  \mathcal{H}^{(id)}_{i, j} ={} & \mathbbm{1}_{\{i \not\in \bm{\Phi}^{(id)}\}}\left[\eta_{c} \mathcal{H}^{(c)}_{i, j} + \eta_{b} \mathcal{H}^{(b)}_{i, j} + \eta_{o} \mathcal{H}^{(o)}_{i, j} + \eta_{a} \mathcal{H}^{(a)}_{i, j}\right], \\
  \mathcal{H}^{(c)}_{i, j} ={} & -\hat{c}_{j}, \\
  \mathcal{H}^{(b)}_{i, j} ={} & \left\|\bm{b}_{i} - \bm{\hat{b}}_{j}\right\|_{1}, \\
  \mathcal{H}^{(o)}_{i, j} ={} & -f_{GIoU}\left(\bm{b}_{i}, \bm{\hat{b}}_{j}\right), \\
  \mathcal{H}^{(a)}_{i, j} ={} & -\left(\frac{\bm{a}^{T}_{i}\bm{\hat{a}}_{j} + \left(\bm{1} - \bm{a}_{i}\right)^{T}\left(\bm{1} - \bm{\hat{a}}_{j}\right)}{N_{a}}\right),
\end{align}
where $\bm{\Phi}^{(id)}$ is a set of ground-truth indices that correspond to $\phi^{(id)}$, $\bm{b}_{i} \in [0, 1]^4$ is a ground truth bounding box normalized with the image size, $\bm{a}_{i} \in \{0, 1\}^{N_{a}}$ is a ground truth action label, $f_{GIoU}\left(\cdot, \cdot\right)$ is a function that calculates generalized IoU~\cite{rezatofighi_cvpr2019}, and $\eta_{\{c, b, o, a\}}$ are the hyper-parameters. The Hungarian algorithm is applied to the matching cost to find the optimal assignment $\hat{\omega}^{(id)} = \argmin_{\omega \in \bm{\Omega}_{N_q}}{\sum_{i=1}^{N_q}{\mathcal{H}^{(id)}_{i,\omega(i)}}}$, where $\bm{\Omega}_{N_q}$ is the set of all possible permutations of $N_q$ elements.

The training loss for individual recognition $\mathcal{L}_{id}$ is calculated between matched ground truths and predictions as follows:
\begin{align}
  \mathcal{L}_{id} ={} & \lambda_{c}\mathcal{L}_{c} + \lambda_{b}\mathcal{L}_{b} + \lambda_{o}\mathcal{L}_{o} + \lambda_{a}\mathcal{L}_{a}, \\
  \mathcal{L}_{c} ={} & \frac{1}{|\bar{\bm{\Phi}}^{(id)}|} \sum_{i=1}^{N_{q}}\left[
        \mathbbm{1}_{\{i \not\in \bm{\Phi}^{(id)}\}}l_{f}\left(\begin{bmatrix}1\end{bmatrix}, \begin{bmatrix}\hat{c}_{\hat{\omega}^{(id)}\left(i\right)}\end{bmatrix}\right) + \mathbbm{1}_{\{i \in \bm{\Phi}^{(id)}\}}l_{f}\left(\begin{bmatrix}0\end{bmatrix}, \begin{bmatrix}\hat{c}_{\hat{\omega}^{(id)}\left(i\right)}\end{bmatrix}\right)\right], \\
  \mathcal{L}_{b} ={} & \frac{1}{|\bar{\bm{\Phi}}^{(id)}|} \sum_{i=1}^{N_{q}} \mathbbm{1}_{\{i \not\in \bm{\Phi}^{(id)}\}}\left\|\bm{b}_i - \bm{\hat{b}}_{\hat{\omega}^{(id)}\left(i\right)}\right\|_{1}, \\
  \mathcal{L}_{o} ={} & \frac{1}{|\bar{\bm{\Phi}}^{(id)}|} \sum_{i=1}^{N_{q}} \mathbbm{1}_{\{i \not\in \bm{\Phi}^{(id)}\}}\left[1 - f_{GIoU}\left(\bm{b}_i, \bm{\hat{b}}_{\hat{\omega}^{(id)}\left(i\right)}\right)\right], \\
  \mathcal{L}_{a} ={} & \frac{1}{|\bar{\bm{\Phi}}^{(id)}|} \sum_{i=1}^{N_{q}}
        \mathbbm{1}_{\{i \not\in \bm{\Phi}^{(id)}\}}l_{f}\left(\bm{a}_{i}, \bm{\hat{a}}_{\hat{\omega}^{(id)}\left(i\right)}\right),
\end{align}
where $\lambda_{\{c, b, o, a\}}$ are hyper-parameters and $l_{f}\left(\cdot, \cdot\right)$ is the element-wise focal loss function~\cite{lin_iccv2017} whose hyper-parameters are described in~\cite{zhou_arxiv2019}.

In our training, the hyper-parameters $\eta_{\{c, b, o, a\}}$ and $\lambda_{\{c, b, o, a\}}$ are set as $\eta_{c} = \lambda_{c} = 1$, $\eta_{b} = \lambda_{b} = 5$, $\eta_{o} = \lambda_{o} = 2$, and $\eta_{a} = \lambda_{a} = 2$.

\section{Implementation Details of Detection Heads}

In our method, all the detection heads are constituted by feed-forward networks with the subsequent sigmoid functions. The details of the detection heads are as follows:
\begin{description}
\item[Person class head]\mbox{}\\
This head has 1 linear layer with the subsequent sigmoid function.
\item[Box head]\mbox{}\\
This head has 3 linear layers with the ReLU activation between the layers and the subsequent sigmoid function. A reference point is added to each corresponding box position before applying the sigmoid function.
\item[Action head]\mbox{}\\
This head has 1 linear layer with the subsequent sigmoid function.
\item[Activity head]\mbox{}\\
This head has 1 linear layer with the subsequent sigmoid function.
\item[Group size head]\mbox{}\\
This head has 3 linear layers with the ReLU activation between the layers and the subsequent sigmoid function.
\item[Member point head]\mbox{}\\
This head has 3 linear layers with the ReLU activation between the layers and the subsequent sigmoid function. $2 \times M$ values are output from the last linear layer and then split into $M$ group member points, where $M$ denotes the maximum group size. A reference point is added to each corresponding group member point before applying the sigmoid function.
\end{description}

\section{Group Annotations in The Volleyball Dataset}
Group annotations are critical components to fully leverage the learning capability of our method. In the evaluation of the Volleyball dataset~\cite{ibrahim_cvpr2016}, we use the original annotation set combined with the extra annotation set provided by Sendo and Ukita~\cite{sendo_mva2019} because the original annotations do not contain group information. The group annotations in the extra set are transferred to the original set by matching bounding boxes from each set with intersection over union (IoU). IoU is first calculated for each pair of a box from the original set and that from the extra set in the same frame. The calculated IoU values are then used as costs for the Hungarian algorithm~\cite{kuhn_naval1955} to match the boxes. If a box from the extra set has a label indicating that the person in the box is involved in an activity, we assign a group member flag to the matched box from the original set.

The players involved in each group activity are defined by Sendo and Ukita~\cite{sendo_mva2019} as follows:
\begin{description}
\item[Pass]\mbox{}\\
Players who are trying an underhand pass independently of whether or not they successfully do it.
\item[Set]\mbox{}\\
A player who is doing an overhand pass and those who will spike the ball whether they are trying or faking.
\item[Spike]\mbox{}\\
Players who are spiking and blocking.
\item[Winpoint]\mbox{}\\
All players in the team  scoring a point. This group activity is observed for a few seconds right after scoring.

\end{description}

\section{Additional Qualitative Analysis}

\begin{figure}	
 \centering
 \begin{subfigure}[t]{1.0\linewidth}
  \centering
  \includegraphics[width=1.0\linewidth]{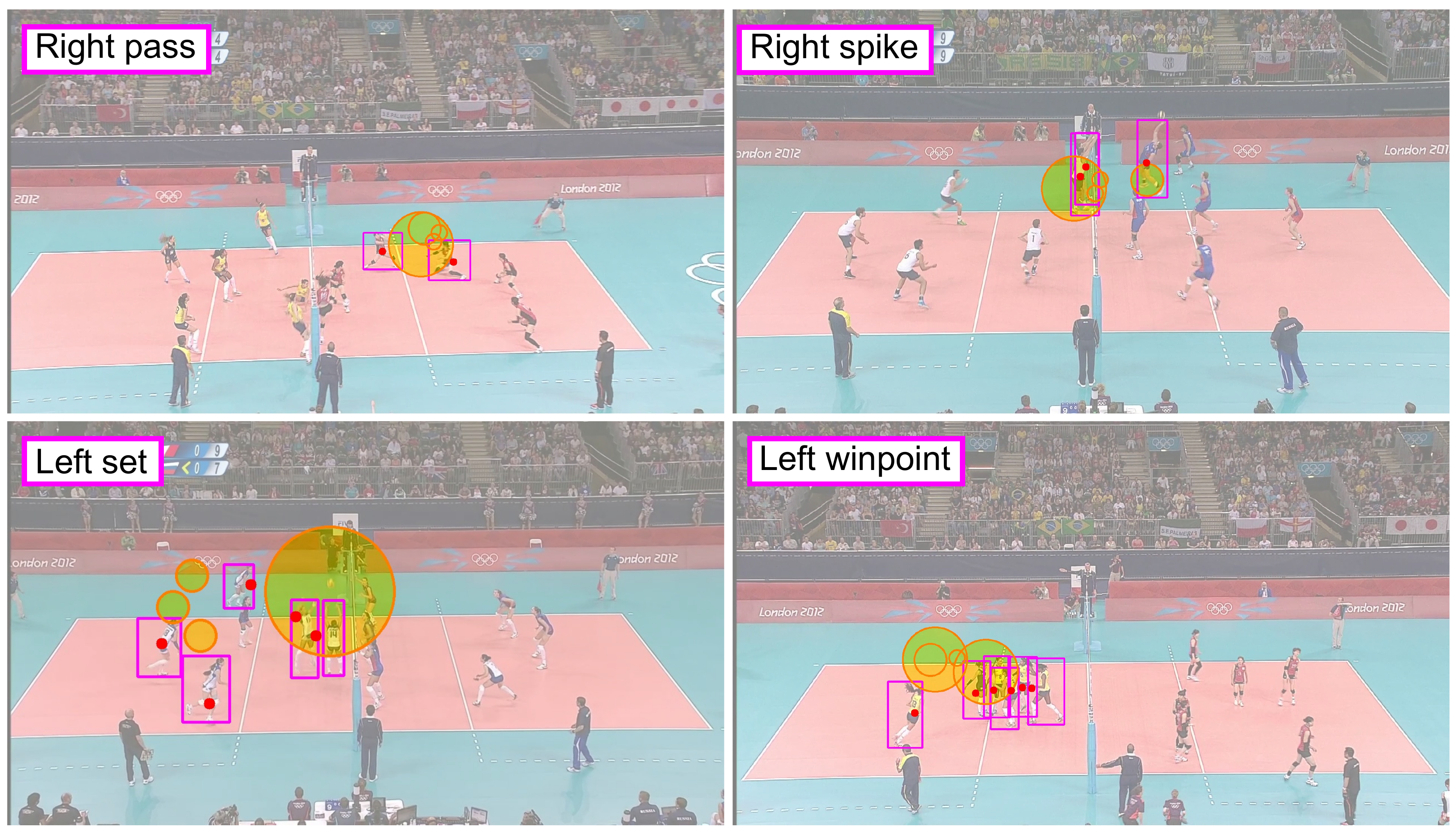}
  \caption{Successful cases.}\label{fig:q_positives}
 \end{subfigure} \\
 \begin{subfigure}[t]{1.0\linewidth}
  \centering
  \includegraphics[width=1.0\linewidth]{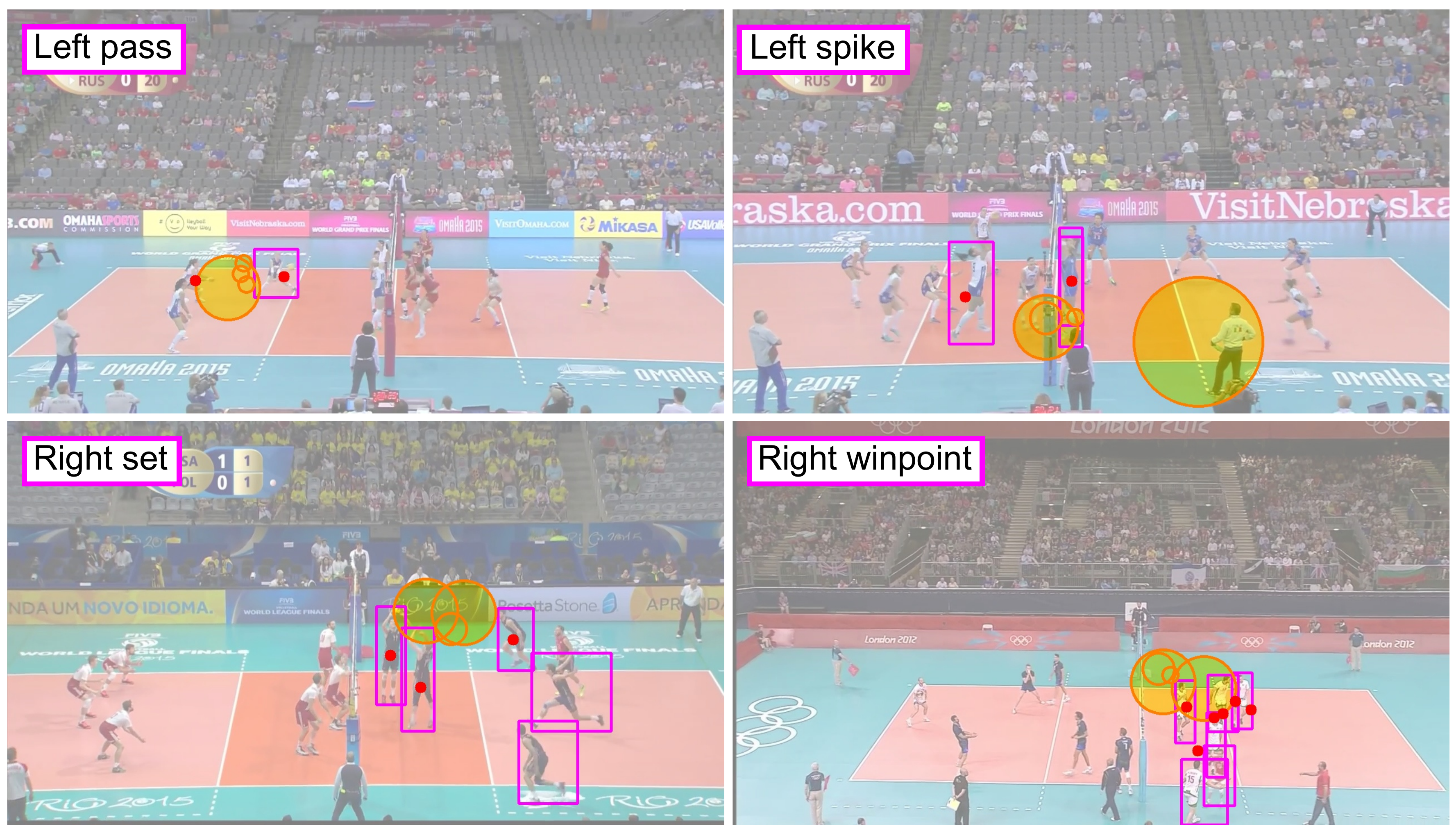}
  \caption{Failure cases.}\label{fig:q_negatives}
 \end{subfigure}
 \caption{Visualization of the social group activity recognition results. The purple bounding boxes, red circles, and yellow circles show the ground truth group members, predicted group member points, and attention locations in the deformable transformer decoder, respectively.}\label{fig:qualitatives}
 \vspace{-1.0ex}
\end{figure}

We further analyze the recognition results qualitatively with our method's success and failure cases on the Volleyball dataset~\cite{ibrahim_cvpr2016}. The results of the successful cases and failure cases are shown in Fig.~\ref{fig:q_positives} and~\ref{fig:q_negatives}, respectively. The purple bounding boxes show the ground truth group members, the red circles show the predicted group member points, and the yellow circles show the attention locations. The small and large yellow circles mean that the locations are in the high and low-resolution feature maps, respectively, offering a rough range of image areas affecting the features used for the predictions.

As seen from the figures, features are successfully aggregated from the areas around the group members in the successful cases, while those are aggregated from the regions around the non-group members, backgrounds, and part of the group members in the failure cases. It is worth noting that our method successfully recognizes social group activities even when one group member is apart from the other members such as the cases of ``Right spike'' and ``Left winpoint'' in Fig.~\ref{fig:q_positives}, demonstrating the effectiveness of our feature aggregation method. In the failure case of ``Left pass'', a non-group member is falsely recognized as a group member probably because the non-group member has the pose of the underhand pass, which is quite similar to the group member. In the failure case of ``Left spike'', a group member cannot be identified due to the occlusion. To correctly identify group members in these cases, long-term temporal context should be leveraged effectively. In the failure cases of ``Right set'' and ``Right winpoint'', the group members are widely distributed especially in the vertical direction. As discussed in the main manuscript, the group member point prediction is designed on the assumption that group members are seen side by side at the same vertical positions in an image. This design might affect the performance of the failure cases. These observations present opportunities for future work.

\end{document}